\documentclass{article}

\usepackage{microtype}
\usepackage{graphicx}
\usepackage{subfigure}
\usepackage{booktabs} 

\usepackage{tabularx}
\usepackage{multirow}
\usepackage{diagbox}
\usepackage{adjustbox}

\usepackage{color}
\usepackage{colortbl}
\usepackage{xcolor}
\usepackage{array}
\definecolor{mygray}{gray}{.9}
\usepackage{subcaption}
\usepackage{wrapfig}

\definecolor{dark}{RGB}{122,149,185}
\definecolor{light}{RGB}{208,228,227}

\usepackage{hyperref}

 \usepackage[accepted]{icml2024}

\usepackage{amsmath}
\usepackage{amssymb}
\usepackage{mathtools}
\usepackage{amsthm}

\usepackage[capitalize,noabbrev]{cleveref}

\theoremstyle{plain}

\theoremstyle{definition}

\theoremstyle{remark}

\usepackage[textsize=tiny]{todonotes}

\icmltitlerunning{IBD: Alleviating Hallucinations in Large Vision-Language Models via Image-Biased Decoding}

\begin{document}

\twocolumn[
\icmltitle{IBD: Alleviating Hallucinations in Large Vision-Language Models via \\Image-Biased Decoding}



\icmlsetsymbol{equal}{*}

\begin{icmlauthorlist}
\icmlauthor{Lanyun Zhu}{sutd}
\icmlauthor{Deyi Ji}{comp}
\icmlauthor{Tianrun Chen}{zju}
\icmlauthor{Peng Xu}{comp}
\icmlauthor{Jieping Ye}{comp}
\icmlauthor{Jun Liu}{sutd}
\end{icmlauthorlist}

\icmlaffiliation{sutd}{Singapore University of Technology and Design}
\icmlaffiliation{comp}{Alibaba Group }
\icmlaffiliation{zju}{Zhejiang University}

\icmlcorrespondingauthor{Lanyun Zhu}{ lanyun\_zhu@mymail.sutd.edu.sg}
\icmlcorrespondingauthor{Jun Liu}{jun\_liu@sutd.edu.sg}

\icmlkeywords{Machine Learning, ICML}

\vskip 0.3in
]



\printAffiliationsAndNotice{}  

\begin{abstract}
Despite achieving rapid developments and with widespread applications, Large Vision-Language Models (LVLMs) confront a serious challenge of being prone to generating hallucinations. An over-reliance on linguistic priors has been identified as a key factor leading to these hallucinations. In this paper, we propose to alleviate this problem by introducing a novel image-biased decoding (IBD) technique. Our method derives the next-token probability distribution by contrasting predictions from a conventional LVLM with those of an image-biased LVLM, thereby amplifying the correct information highly correlated with image content while mitigating the hallucinatory errors caused by excessive dependence on text. We further conduct a comprehensive statistical analysis to validate the reliability of our method, and design an adaptive adjustment strategy to achieve robust and flexible handling under varying conditions. Experimental results across multiple evaluation metrics verify that our method, despite not requiring additional training data and only with a minimal increase in model parameters, can significantly reduce hallucinations in LVLMs and enhance the truthfulness of the generated response. 
\end{abstract}

\section{Introduction}
The rapid development of large language models (LLMs) \cite{touvron2023llama, brown2020language} in recent years marks a significant step toward achieving artificial general intelligence (AGI). Large vision-language models (LVLMs) \cite{liu2023visual, wu2023visual, wei2023vary, wang2023cogvlm} have further extended the capabilities of LLMs to the visual domain, demonstrating impressive performance across various tasks such as image captioning \cite{li2023blip, wang2023caption} and grounding \cite{rasheed2023glamm, you2023ferret}. Despite their notable successes, LVLMs still encounter numerous challenges that impede their real-world utility. Among these challenges, the phenomenon of hallucinations is particularly critical, with its frequent occurrence prompting extensive worries about the safety and stability of artificial intelligence systems.

\begin{figure}
    \centering
    \includegraphics[width=0.9\linewidth]{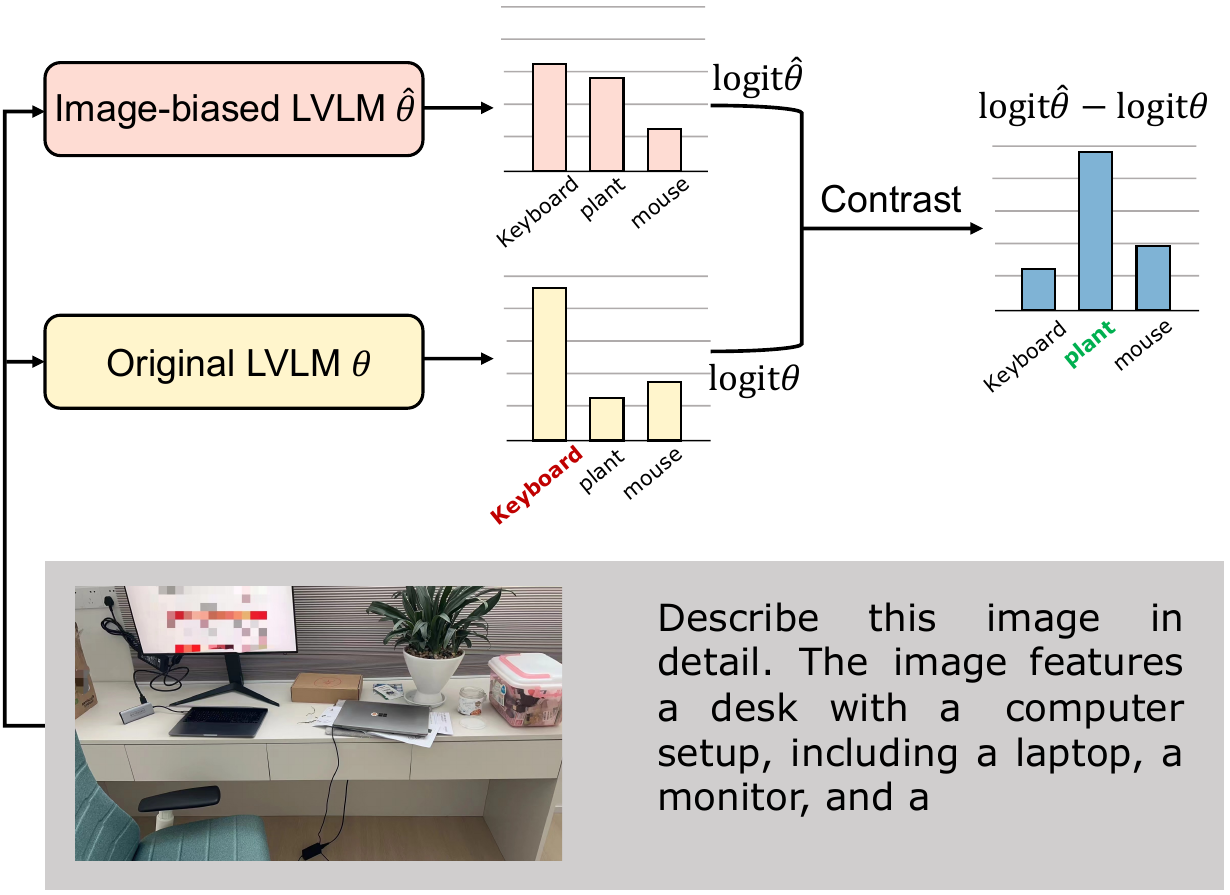}
    \vspace{-0.5\baselineskip}
    \caption{An illustrative example of our method. Texts highlighted in {\color{red} red} and {\color{green} green} indicate erroneous prediction and correct prediction generated by the original LVLM and contrastive results, respectively.}
    \label{fig_intro}
\end{figure}

In the field of vision-language models, the term `hallucination' denotes the generation of information by the model that is either unrelated or incorrect with respect to the given text and image inputs. Recent studies \cite{holtzman2019curious} suggest that the standard pretraining mechanism of LLMs, which targets the maximum-likelihood-based next token prediction, is a significant contributor to the occurrence of hallucinations. This mechanism may cause models to depend excessively on superficial patterns found in the training data rather than developing a substantial understanding of fundamental logical principles \cite{ji2023survey}. Specifically for the vision-language models, it has been observed that LVLMs finetuned from language models tend to depend excessively on linguistic priors during the autoregressive text generation process \cite{han2022visual, li2023evaluating, leng2023mitigating}. As a result, responses are often simply continued based on the text generated from the previous time steps, neglecting to draw logical inferences from the actual visual content within the input image. This issue of language over-reliance is particularly severe in the generation of long text, with empirical observations \cite{zhou2023analyzing} indicating that the later segments of the generated response are more susceptible to hallucinations.

In this work, we aim to address the issue of hallucinations in LVLMs by proposing a novel contrastive decoding technique \cite{li2022contrastive} that is designed based on the underlying cause of these hallucinations to extract accurate information. Our proposed method involves computing a more reliable next-token probability distribution by contrasting the predictions of the original model with those of an image-biased model that focuses more on the image information. The image-biased model is created by modifying the attention weight matrix structure within the original model without altering its parameters. This approach emphasizes the knowledge of the image-biased model and diminishes that of the original model, which may be text-biased, thus encouraging the extraction of correct content while suppressing hallucinations resulting from textual over-reliance. An illustrative example of our method is presented in Fig.\ref{fig_intro}. The original model erroneously predicts `keyboard' as the next token, a term often associated with `computer' in linguistic priors but absent in the input image. Upon using the image-biased model, the correct token `plant' receives a significantly higher probability boost. Consequently, the contrast between these two predictions can serve as an effective indicator to select the accurate tokens and mitigate hallucinations.

To ensure the reliability of our method, we conduct a statistical analysis to verify the alignment between the tokens receiving the highest probability boost and the correct tokens devoid of hallucinations. The results reveal a significant correlation in certain conditions, demonstrating the credibility of our method. Furthermore, we identify potential scenarios where our method may fail and, in response, develop a dynamic adjustment mechanism to address such issues. By incorporating these designs, our method, named \textbf{I}mage-\textbf{B}iased \textbf{D}ecoding (IBD), demonstrates notable effectiveness in mitigating hallucinations. The benefits of IBD can be summarized in three key aspects: (1) \textbf{Minimal Overhead}. IBD operates with low extra parameter and data costs, distinguishing it from previous methods like \cite{yu2023hallucidoctor, liu2023mitigating}, which necessitate extra training data, or \cite{yin2023woodpecker, zhou2023analyzing} which demand a considerable expansion of model parameters. (2) \textbf{Comprehensive Processing Capability}. Unlike earlier contrastive decoding strategies \cite{li2022contrastive, leng2023mitigating}, IBD is designed based on an in-depth statistical analysis that takes into account the unique properties of different vocabulary types, allowing it to adaptively handle a diverse range of situations flexibly. (3) \textbf{Superior Performance}. Extensive validation experiments are conducted to benchmark IBD against other advanced methods. The leading results across multiple metrics demonstrate the high effectiveness of IBD.

\section{Related Work}
\noindent \textbf{Large Vision-Language Models. }The introduction of large language models (LLMs) \cite{touvron2023llama, brown2020language} has marked the advent of a new era in artificial intelligence. Initially, LLMs were limited to text processing. Recent advancements have expanded their capabilities to process images, leading to the development of Large Vision-Language Models (LVLMs) \cite{dai2023instructblip, liu2023visual, zhu2023minigpt, bai2023qwen} in a multi-modal manner. These LVLMs have achieved impressive and generalizable performance across multiple tasks, such as image captioning \cite{li2023blip}, visual question answering \cite{zhang2023pmc}, object detection \cite{xu2023pixel} and image segmentation \cite{lai2023lisa, zhu2023llafs}. Despite their successes, similar to LLMs, LVLMs are prone to generating hallucinations \cite{li2023evaluating}, which significantly hampers their robust application. Our research aims to address the issue of hallucinations in LVLMs to improve their practical utility and reliability.

\noindent \textbf{Hallucinations in VLMs and LVLMs. } Hallucination refers to the problem that a model generates information that is either irrelevant or incorrect with respect to the given context. This issue has attracted increased attention within the domain of VLMs \cite{kim2023exposing, rohrbach2018object}. Recent researchers \cite{wang2023evaluation, zhao2023beyond, ben2023mocha} have started to focus intensively on the hallucination phenomenon in LVLMs. Some approaches \cite{yu2023hallucidoctor, liu2023mitigating, wang2023vigc}  attempt to mitigate hallucinations by fine-tuning the LVLMs on fine-grained datasets, while others \cite{wang2023mitigating, yin2023woodpecker, zhou2023analyzing} propose to use supplementary networks to identify hallucinated objects or to rewrite the output text from LVLMs. However, these techniques typically necessitate a substantial number of additional model parameters or training data. Our method differs from these approaches by focusing only on refining the model's decoding process, thereby effectively reducing hallucinations without requiring too many additional parameters or data. 

\noindent \textbf{Decoding Method. }Decoding method determines the generation of text tokens at each time step within language models. Traditional decoding strategies such as beam search \cite{boulanger2013audio}, top-k decoding \cite{fan2018hierarchical}, and sampling methods \cite{holtzman2019curious}, despite their widespread use, are prone to producing hallucinatory content. Recent research \cite{li2022contrastive, chuang2023dola, leng2023mitigating, huang2023opera} has made attempts to address this issue by proposing better decoding methods. For instance, \cite{leng2023mitigating} uses contrastive decoding in LVLMs; however, this technique relies on randomly generated noise as input, which can introduce uncontrollable and unstable informational perturbations. In contrast to these existing approaches, our method delves into the fundamental causes of hallucinations in LVLMs and tailors a comprehensive method in consideration of varying textual words, leading to an efficient and robust solution for handling a wide array of scenarios. Table.\ref{chair_result} presents the comparison results, where our approach demonstrates superior performance than other decoding methods. 

\section{Method}
Recent Large Vision-Language Models (LVLMs) typically utilize a visual image $v$ and a text $t$ as inputs to generate a text response. The process begins with passing the input image $v$ through a visual encoder, followed by a feature space projector to generate a set of visual tokens. These visual tokens are subsequently concatenated with the input text tokens to serve as the input for the LVLM's language model. Within the framework, the LVLM generates the response $y=\{y_{1}, y_{2}, ..., y_{N_{y}}\}$ with $N_{y}$ time steps in an auto-regressive manner, with the probability distribution of generating token $y_{i}$ at the $i$-th time step being represented as $p_{\theta}\left(y_{i}|v, t, y_{<i}\right) = {\rm Softmax}\left({\rm logit}_{\theta}\left(y_{i}|v, t, y_{<i}\right)\right)$, where $\theta$ refers to the LVLM model, and $y_{<i}=\{y_{1}, y_{2}, ..., y_{i-1}\}$ denotes the response generated from the previous time steps.

In contrast to vanilla LVLMs, our method does not directly utilize the probability distribution obtained through the aforementioned process for prediction. Instead, to alleviate hallucinations, we obtain next-token prediction by contrasting the ordinary prediction with an image-biased prediction. Specifically, alongside the existing model $\theta$, we introduce a new model $\hat{\theta}$. Different from $\theta$, which theoretically balances the use of both image $v$ and text $\{t, y_{<i}\}$ for prediction, $\hat{\theta}$ exhibits a higher dependency and places greater emphasis on the information contained within the image $v$. With the prediction logits ${\rm logit}_{\hat{\theta}}\left(y_{i}|v, t, y_{<i}\right)$ obtained using $\hat{\theta}$, we compute the next-token-probability by:
\begin{equation}\label{cd_pre}
p\left(y_{i}|y_{<i}\right) = {\rm Softmax}\left(\mathcal{L}_{CD}\right),
\end{equation}
\begin{equation}\label{cd}
    \mathcal{L}_{CD} = {\rm logit}_{\hat{\theta}}\left(y_{i}|v, t, y_{<i}\right) - {\rm logit}_{\theta}\left(y_{i}|v, t, y_{<i}\right),
\end{equation}
Where $\mathcal{L}_{CD}$ is named as CD score. The motivation for introducing $\hat{\theta}$ and the contrastive decoding method is based on the hallucination characteristics observed in LVLMs. Previous studies \cite{han2022visual, li2023evaluating} have identified the excessive dependence on the language model's linguistic priors as a critical contributor to LVLM's hallucinations. In particular, it has been observed that LVLMs can occasionally disregard input image's information but treat the next-token prediction as a purely text-based continuation task. Such an approach can lead to the generation of responses that are unrelated to the input image and are therefore text-biased hallucinations. Based on these observations, we propose a hypothesis that, using the model $\hat{\theta}$ that focuses more on input image information when making predictions in each time step, the token with the highest probability boost from $\theta$ to $\hat{\theta}$ among the candidate pool is more likely to be the correct token without text-biased hallucinations. We provide a detailed description of model $\hat{\theta}$ in Sec.\ref{sec:image_biased}; and in Sec.\ref{sec:hyp_test}, we examine our hypothesis regarding the differences in results when utilizing $\theta$ and $\hat{\theta}$.

\subsection{Image-biased Model} \label{sec:image_biased}
We employ a simple yet effective approach to construct the image-biased model $\hat{\theta}$, by merely adjusting the attention weight matrices within the vanilla model $\theta$. Within each attention weight matrix of LVLM's language model, we apply an amplification coefficient to the image tokens, thereby increasing the attention's focus on visual information while minimizing its interaction with textual features. This amplification coefficient $c$ is a pre-defined numerical value, which is conditionally added to the attention's query-key multiplication result when the key corresponds to an image token. To be specific, in the $i$-th time step for generating the text response, we denote the tokens for the $l$-th layer of the LVLM as $\mathbf{T}^{l}=\{\mathbf{T}^{l}_{v},\mathbf{T}^{l}_{o}$\}, where $\mathbf{T}_{v}^{l}=\{\mathbf{T}_{v, k}^{l}\}_{k=1}^{N_{v}}$ and $\{\mathbf{T}_{o, k}^{l}\}_{k=1}^{N_{o}}$ refer to the tokens of the input image $v$ and other components (input text, previously generated response, padding tokens), respectively. After that, we use the following formulation to compute each item $W^{l}_{m,n}$ of the attention's weight matrix $W^{l}$: 
\begin{equation} \label{weight}
\begin{aligned}
&W^{l}_{m,n} = {\rm Softmax}\left(\frac{Q^{l}_{m}(K^{l}_{n})^{\rm T}}{\sqrt{D}}+c_{m,n} + M\right),\\
&c_{m,n}=\epsilon\quad {\rm if} \quad K^{l}_{n}\gets \mathbf{T}_{v}^{l} \quad {\rm else} \quad 0,
\end{aligned}
\end{equation}
where $Q^{l}$ and $K^{l}$ refer to the query and key tokens derived from $\mathbf{T}^{l}$, $M$ represents the casual mask, and $D$ denotes the feature dimension. $\quad K^{l}_{n}\gets \mathbf{T}_{v}^{l}$ represents that the $n$-the key token $K^{l}_{n}$ corresponds to an image token. Utilizing the method, we encourage the attention mechanism to place greater focus on image information, consequently making the constructed model $\hat{\theta}$ to become image-biased. Note that $\hat{\theta}$ shares the same weight parameters with $\theta$, thus preventing the additional memory usage to load other parameters during the inference stage. In Sec.\ref{ablation}, we will discuss the detailed settings for the hyperparameter $\epsilon$ in Eq.\ref{weight}.

\begin{figure}[t]
    \centering
    \hspace{-0.07\linewidth}
    \subfigure[Content Words]{
        \label{fig:sub1}
        \centering
        \includegraphics[width=0.51\linewidth]{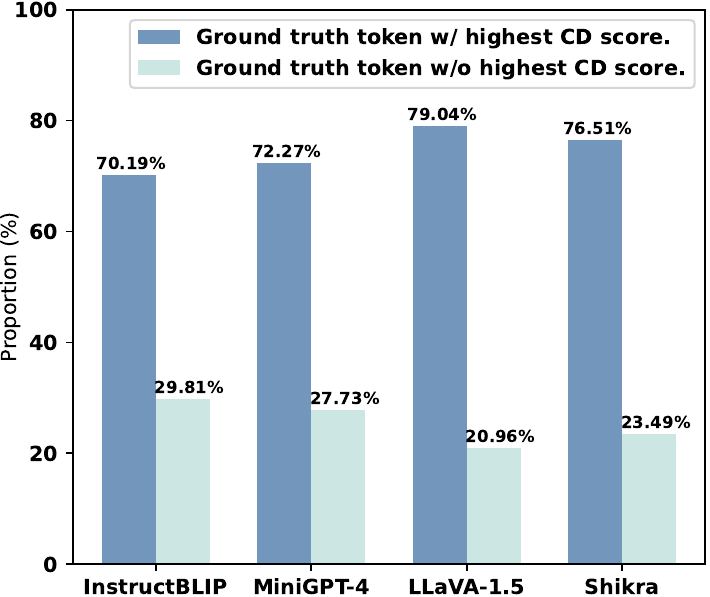}
        }
    \hspace{-0.05\linewidth}
    \subfigure[Function Words]{
        \label{fig:sub2}
        \centering
        \includegraphics[width=0.51\linewidth]{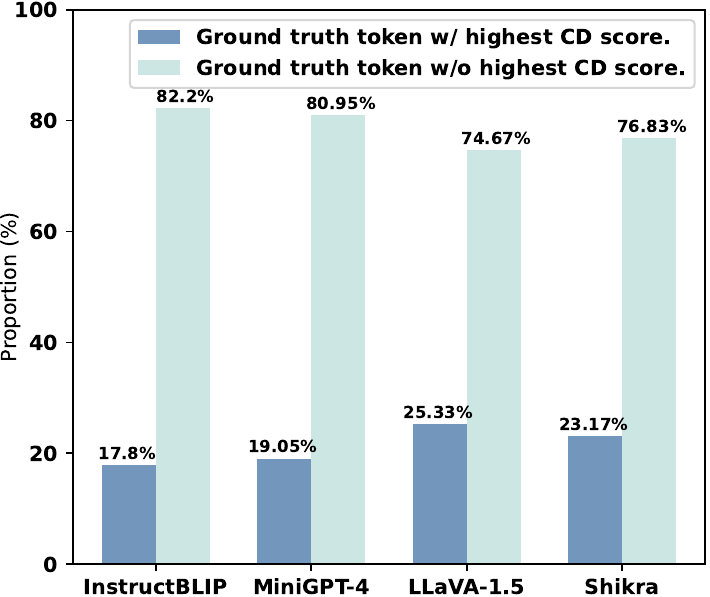}
        }
    \vspace{-0.5\baselineskip}
    \caption{(a) and (b) respectively present the statistical results for content words and function words in the COCO Caption dataset. \textcolor{dark}{\textbf{Dark}} bars represent the proportion of ground tokens with the highest CD score among all candidate tokens, while \textcolor{dark}{light} bars represent the proportion of ground tokens without the highest CD score. The statistical results on four LVLMs are reported, including InstructBLIP, MiniGPT-4, LLaVA-1.5 and Shikra.}
    \label{fig_content_function}
\end{figure}

\begin{figure}[t]
    \centering
    \hspace{-0.07\linewidth}
    \subfigure[InstructBLIP]{
         \label{blip_dis}
        \centering
        \includegraphics[width=0.51\linewidth]{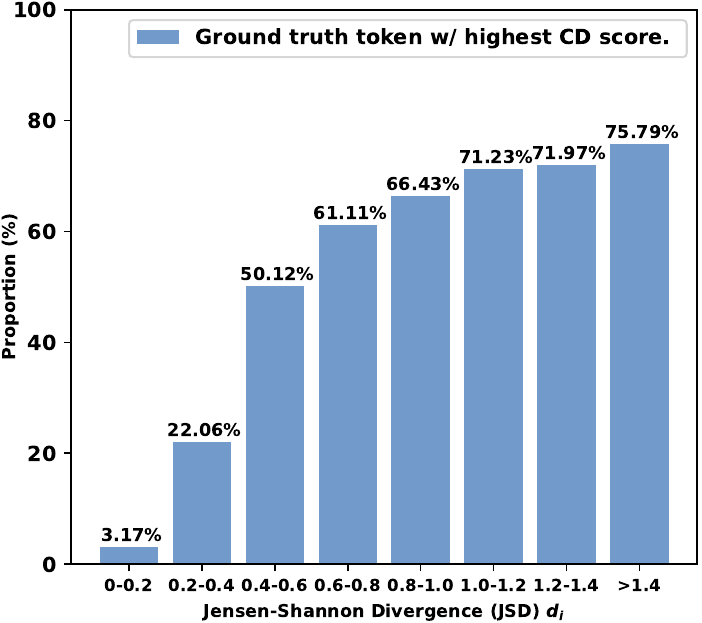}
        }
    \hspace{-0.05\linewidth}
    \subfigure[MiniGPT-4]{
        \label{mini_dis}
        \centering
        \includegraphics[width=0.51\linewidth]{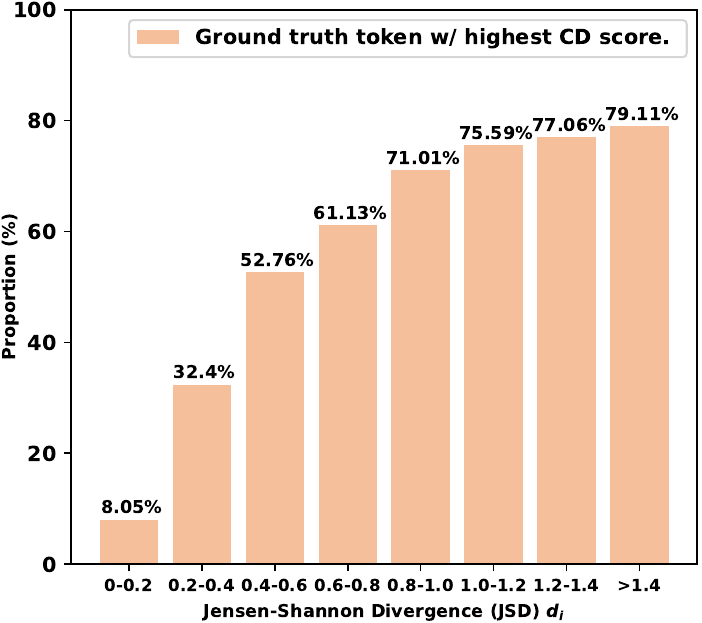}
        }
    \vspace{-0.5\baselineskip}
    \hspace{-0.07\linewidth}
    \subfigure[LLaVA-1.5]{
        \label{llava_dis}
        \centering
        \includegraphics[width=0.51\linewidth]{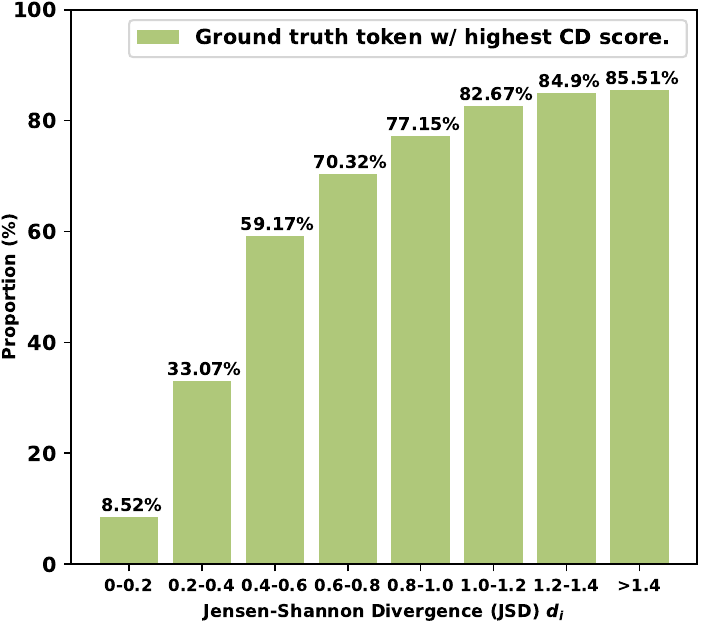}
        }
    \hspace{-0.05\linewidth}
    \subfigure[Shikra]{
        \label{shikra_dis}
        \centering
        \includegraphics[width=0.51\linewidth]{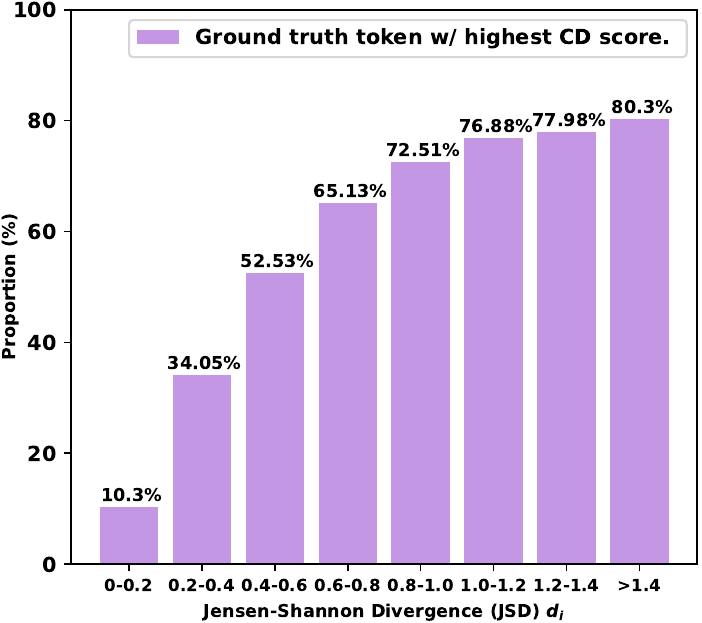}
        }
    \caption{Statistical results to illustrate the relationship between the prediction similarity of $\theta$ and $\hat{\theta}$ and the proportion of ground tokens having the maximum CD score among candidate tokens. X-axis denotes the range of Jensen-Shannon divergence (JSD) $d_{i}$ between the prediction results from $\theta$ and $\hat{\theta}$. $d_{i}$ is scaled by $1.5\times10^{4}$. A higher $d_{i}$ indicates lower similarity. Y-axis represents, for all time steps with its $d_{i}$ falling into each range, the proportion of time steps where the ground truth token has the highest CD score among all candidate tokens.}
    \label{dis}
\end{figure}

\subsection{Hypothesis Testing}\label{sec:hyp_test}
Based on $\hat{\theta}$, we evaluate the hypothesis presented above, i.e., in each time step, candidate token with the highest probability boost from $\theta$ to $\hat{\theta}$ (CD score $\mathcal{L}_{CD}$ as Eq.\ref{cd}) is more likely to be the correct response without text-biased hallucinations. We conduct experiments on COCO datasets to test this hypothesis, and we find the following two patterns:

\textbf{Difference for Content Words and Function W}\textbf{ords. }
Our first finding is that the proposed hypothesis holds true for content words but does not apply to function words. In linguistics, content words refer to words including rich information like most of the nouns and adjectives; function words, on the other hand, refer to other words for forming sentence structures but with less informational content, such as `a', `the', `to', `and', etc. We employ ChatGPT to categorize each word in the COCO captions into these two types. For each time step of a caption text, we compute the CD scores $\mathcal{L}_{CD}$ using Eq.\ref{cd}, and assess whether the ground truth token attains the highest CD score among all candidate tokens\footnote{Candidate tokens include one ground truth token and the other hallucinatory candidate tokens}. This assessment is conducted across all caption words with the ground truth classified as either content or function words, followed by a statistical analysis shown in Fig.\ref{fig_content_function}. Our findings indicate that, for content words, the ground truth tokens frequently receive the highest scores, thus validating our hypothesis for these terms. In contrast, for function words, the hypothesis is not supported, and the ground truth tokens frequently score lower than the other hallucinatory candidate tokens.
We attribute this pattern to the LVLM's reliance on different types of information when predicting different words. Content words, often closely associated with image content, benefit from a model's increased focus on image data. Conversely, function words, devoid of visual information and heavily tied to grammatical constructs, depend more on the linguistic priors within the LLM. Consequently, using an image-biased $\hat{\theta}$ may increase the probability of predicting content tokens while reducing the probability of predicting function tokens, thus leading to higher CD scores for the content ground truth tokens and lower scores for the function ground truth tokens.

\noindent \textbf{Impact of Prediction Similarity between $\theta$ and $\hat{\theta}$.} Another interesting finding is that the proposed hypothesis may not hold when the predictions $p_{\theta}\left(y_{i}|v, t, y_{<i}\right)$ from $\theta$ and $p_{\hat{\theta}}\left(y_{i}|v, t, y_{<i}\right)$ from $\hat{\theta}$ are too similar. Specifically, we use $d_{i}$, calculated as the Jensen-Shannon divergence (JSD), to quantify the similarity of the predictions by $d_{i} = {\rm JSD}\left(p_{\theta}\left(y_{i}|v, t, y_{<i}\right) || p_{\hat{\theta}}\left(y_{i}|v, t, y_{<i}\right)\right)$. Our statistical results, as shown in Fig.\ref{dis}, suggest that a reduction in $d_{i}$ correlates with a greater probability that the candidate token with the highest CD score is NOT the actual ground truth in the $i$-th time step. When the prediction similarity between $\theta$ and $\hat{\theta}$ is too high, the CD score, computed as the prediction difference, would be low in value and likely to be influenced by noise, thus leading to a reduced accuracy by using it for identifying the correct token. This finding is consistent with the observations reported in the previous research \cite{li2022contrastive, chuang2023dola}.

A low value of $d_{i}$ could be attributed to two reasons: (1) the prediction from $\theta$ for $y_{i}$ might already be heavily image-dependent, and thus, further enhancing image focus in $\hat{\theta}$ does not lead to a significant change in the prediction outcome; (2) the model's hallucinations by over-relying on text coincidentally aligns with the actual content within the image, so the model is capable of achieving similar predictions for $y_{i}$ when either over-relying on textual information in $\theta$ or image in $\hat{\theta}$. Note that we refer to reason (2) as `benign text-biased hallucination. In fact, this phenomena is common in LVLMs, as the LVLM's text-biased hallucinations tend to be biased toward the learned associations from the training set \cite{zhou2023analyzing}. These associations may also likely be present in the test images. For instance, upon predicting the word `computer', an LVLM might frequently predict `keyboard' subsequently without relying on image information, because `computer' and `keyboard' are commonly seen together in the training corpus. If a keyboard is indeed in the test image that also contains a computer, this sort of benign text-biased hallucinations will not lead to an erroneous prediction.

\subsection{Dynamic Adjustment For Decoding}\label{dynamic_adjustment}
The above analysis reveals that an over-reliance on textual information is not always detrimental. In the case of predicting function words, a dependence on language priors is appropriate, whereas a bias toward images is detrimental. Moreover, when predictions from $\theta$ and $\hat{\theta}$ are closely aligned, there is no substantial difference between text-reliant and image-reliant outcomes. Consequently, it is impractical to employ CD scores for the prediction of such terms. Through intuitive analysis, we assert that even when using the vanilla maximum likelihood decoding (greedy decoding) method, these terms are less prone to suffer from errors caused by text-biased hallucinations. This is because function words are generally easy to predict. While terms with similar predictions from $\theta$ and $\hat{\theta}$—either already emphasizing image information when predicting or exhibiting benign text-biased hallucinations—are less likely to be the erroneous information caused by the over-reliance on text.

Building upon the above discussion, we introduce a dynamic mechanism for token prediction, which adaptively adjusts the balance between the traditional maximum likelihood decoding method and the proposed image-biased contrastive decoding. This self-adaptive adjustment is implemented according to the state of the current decoding step. 
Formally,
\begin{equation}\label{dynamic}
\begin{aligned}
&y_{i} \sim {\rm Softmax}\left({\rm logit}_{\theta}\left(y_{i}|v, t, y_{<i}\right) + \alpha \cdot I\cdot \mathcal{L}_{CD}\right),\\
&{\rm where}\; I = {\rm Min}\{I_{sim}, I_{con}\},
\end{aligned}
\end{equation}
where $\alpha$ is a scaling factor, $I_{sim}$ and $I_{con}$ are two indicators. $I_{sim}$ reflects the difference between the predictions from $\theta$ and $\hat{\theta}$, and $I_{con}$ reflects the probability that the token in the current decoding time step is a content word. In practice, we compute $I_{sim}$ as the Jensen-Shannon divergence by:
\begin{equation}\label{i_sim}
    I_{sim} = {\rm JSD}\left(p_{\theta}\left(y_{i}|v, t, y_{<i}\right) || p_{\hat{\theta}}\left(y_{i}|v, t, y_{<i}\right)\right).
\end{equation}
To calculate $I_{con}$, one might consider employing an existing POS tagging tool to distinguish whether a predicted token is a content or function word. This method, however, is prone to semantic ambiguities due to subword tokenization. For instance, the token `on' generated in a time step might be part of the content word `onion'. Directly labeling `on' as a function word could therefore introduce inaccuracies. To address this issue, we instead adopt an implicit approach. Prior research \cite{chuang2023dola} has shown that when employing an early exit strategy \cite{teerapittayanon2016branchynet} in language models to predict function words, the LLM tends to determine the token to generate within the middle layers, and keeping the predictions almost unchanged in the subsequent higher layers. In contrast, when predicting content words that are more complex, the LLM continues to change its predictions in the last few layers. Our empirical investigation, detailed in the appendix, demonstrates that this behavior also applies to LVLMs and subwords that constitute content words. Levering these findings, we employ the same early exiting method as \cite{chuang2023dola} to get a prediction $\widetilde{p}_{\theta}\left(y_{i}|v, t, y_{<i}\right)$ from the LVLM's middle layer (the 24-th layer in our setting). We then define $I_{con}$ as the distance between the predictions from the final layer and those from the intermediate layer, which is formally calculated as:
\begin{equation}\label{i_fun}
    I_{con} = {\rm JSD}\left(p_{\theta}\left(y_{i}|v, t, y_{<i}\right) || \widetilde{p}_{\theta}\left(y_{i}|v, t, y_{<i}\right)\right).
\end{equation}
$I_{sim}$ and $I_{con}$ are used in Eq.\ref{dynamic} to complete the prediction process. This dynamic approach allows the model to rely more on maximum likelihood when predicting function words and words that elicit similar predictions from $\theta$ and $\hat{\theta}$, thereby mitigating the negative impact that may arise from a reduced ability of the CD score to accurately identify the correct tokens in certain instances. This harmonization of multiple strategies ensures that the model's performance is not unduly hindered by potential weakness in any single approach, particularly when dealing with some specific types of words as we have discussed before.

\subsection{Full Method} \label{full_method}
Building upon the approaches illustrated above, we further advance our method by introducing two minor yet effective improvements. The first improvement is to finetune the image-biased model $\hat{\theta}$. In our method, $\hat{\theta}$ is derived by modifying the attention weight matrices without altering the parameter weights. However, this modification on model structures might introduce noise, which in turn could impact the reliability of the CD score. To mitigate the issue, we finetune $\hat{\theta}$ on the COCO caption dataset to adapt it better to the modified attention architecture. Note that we do not apply the frequently-used LoRA for finetuning, which requires an excessive number of additional parameters. Instead, we utilize the more lightweight prompt tuning technique by adding a small set of learnable prompts $P$ to the LLM's inputs. With $P$ obtained through finetuning, the output of $\hat{\theta}$ at the $i$-th time step can be formulated as ${\rm logit}_{\hat{\theta}}\left(y_{i}|P, v, t, y_{<i}\right)$. The second improvement draws inspiration from previous methods \cite{li2022contrastive} by employing an adaptive plausibility constraint, which is realized by selectively considering only a portion of tokens that possess high enough output probabilities as the candidates for prediction. This approach helps to prevent the frequent occurrence of false positives and false negatives in contrastive decoding methods \cite{li2022contrastive, chuang2023dola}. Incorporating these improvements, the full method to predict a token in a time step $i$ is written as:
\begin{equation}\label{full}
\begin{aligned}
y_{i} \sim {\rm Softmax}\left({\rm logit}_{\theta}\left(y_{i}|v, t, y_{<i}\right) + \alpha \cdot I\cdot \mathcal{L}_{CD}\right),&\\
{\rm subject\ to}\ y_{i} \in \mathcal{V}_{head}\left(y_{<i}\right),&
\end{aligned}
\end{equation}
\begin{equation}\label{condition}
\begin{aligned}
{\rm where}\;  &\mathcal{L}_{CD} = {\rm logit}_{\hat{\theta}}\left(y_{i}|P, v, t, y_{<i}\right) - {\rm logit}_{\theta}\left(y_{i}|v, t, y_{<i}\right),\\
& I = {\rm Min}\{I_{sim}\left({\rm Eq.\ref{i_sim}}\right),\; I_{con}\left({\rm Eq.\ref{i_fun}}\right)\},\\
& \begin{aligned}&\mathcal{V}_{\mathrm{head}}\left(y_{<i}\right)=\{y_{i}\in\mathcal{V}:\\&\qquad \qquad p_{\theta}\left(y_{i}| v,t,y_{<i}\right)\geq\beta\max_{w}p_{\theta}\left(w| v,t,y_{<i}\right)\},\end{aligned}
\end{aligned}
\end{equation}
where $\mathcal{V}$ is LVLM's output vocabulary and $\beta$ is a hyper-parameter between 0 to 1.

\begin{figure}
    \centering
    \includegraphics[width=1\linewidth]{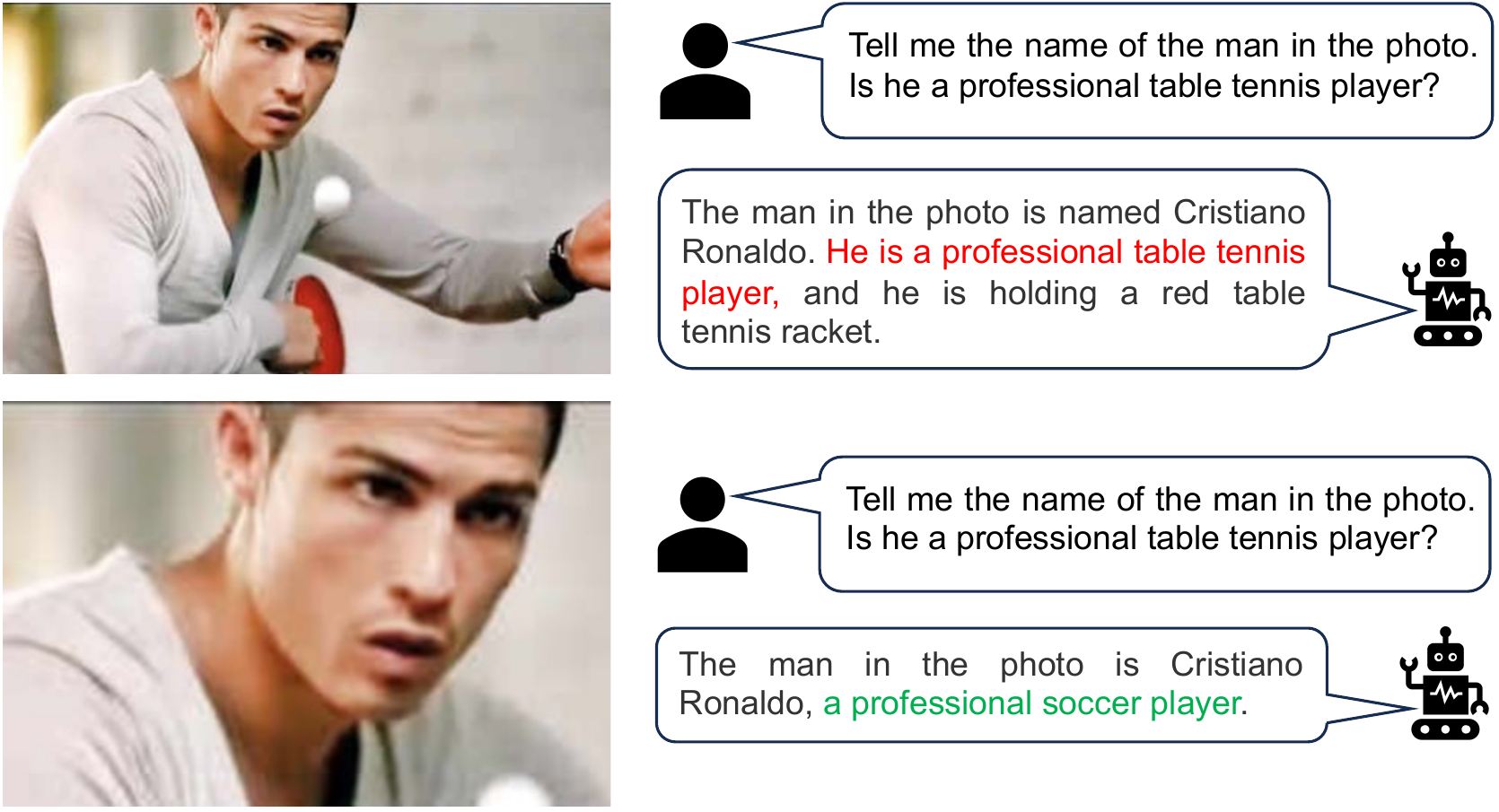}
    \vspace{-0.5\baselineskip}
    \caption{An example to show the problem of image-biased hallucinations in LVLMs. Texts highlighted in {\color{red} red} and {\color{green} green} indicate erroneous information and correct information generated by LLaVA-1.5, respectively.}
    \label{image_h}
\end{figure}

\subsection{Discussion: Are there Image-biased Hallucinations?}\label{discuss_image_biased}
This study builds on the existing understanding that hallucinations in LVLMs typically stem from an over-reliance on textual information. A question that naturally arises is whether LVLM's hallucinations exclusively arise from text dependency, and if some can also be attributed to the excessive reliance on input image. An example shown in Fig.\ref{image_h} demonstrates that such image-biased hallucinations indeed exist. In the example, a photo showing the renowned soccer star Cristiano Ronaldo playing table tennis is presented to an LVLM with the question about whether the man is a professional table tennis player. The LVLM responses incorrectly. Subsequently, the image is cropped to exclude the table tennis racket, and the modified image is resubmitted to the LVLM, which in turn provides the correct response. This result suggests that the LVLM's attention to the paddle region impairs its judgment, thereby validating the existence of image-biased hallucinations.

Our empirical observation suggests that such image-biased hallucinations frequently occur when there is an inconsistency between the visual content and the language model's world knowledge. For example, as shown in Fig.\ref{image_h}, determining if the man in the image is a professional table tennis player relies on the language model's knowledge that Cristiano Ronaldo is a footballer, a fact at odds with the presence of a table tennis racket for another sport in this image. Consequently, overreliance on the visual cue of the racket, without proper integration of the model's world knowledge, may lead to an incorrect inference that the man is a table tennis professional. Such misjudgments, however, are uncommon in the mainstream evaluation frameworks for LVLMs. These frameworks typically encompass tasks such as image captioning, which necessitates a limited scope of world knowledge and focus more on the direct descriptions of visual content, and question-answering tasks that typically process straightforward questions without excessive interference. In these cases, the visual content is less prone to conflicting with the language model's world knowledge, thereby making image-biased hallucinations a rarer phenomenon as demonstrated by our empirical analysis presented in Appendix.

Given the insights obtained from our analyses, and taking into account the characteristics of the existing evaluation frameworks, we have chosen not to create targeted solutions for the image-biased hallucinations. Instead, we propose the development of a more comprehensive assessment method for image-biased hallucinations and the corresponding solutions as a topic for future research.

\vspace{-0.25\baselineskip}
\section{Experiments}
\vspace{-0.25\baselineskip}
\subsection{Settings}
\vspace{-0.25\baselineskip}
\noindent \textbf{Implementation Details.} We set $\epsilon$ in Eq.\ref{weight} to 2, $\alpha$ in Eq.\ref{dynamic} and Eq.\ref{full} to $1.5\times10^{4}$, and $\beta$ in Eq.\ref{condition} to 0.1.$\widetilde{p}_{\theta}\left(y_{i}|v, t, y_{<i}\right)$ in Eq.\ref{i_fun} for computing $I_{con}$ is derived from LVLM's 24-th layer. We present a detailed ablation study for these hyper-parameters in Sec.\ref{ablation} and Appendix. 

\noindent \textbf{Baseline Models.} Following previous research \cite{huang2023opera}, we conduct performance validation and comparison across four mainstream large vision-language models, including InstructBLIP \cite{dai2023instructblip}, MiniGPT-4 \cite{zhu2023minigpt}, LLaVA-1.5 \cite{liu2023visual} and Shikra \cite{chen2023shikra}. All these models utilize an LLM with 7 billion parameters.

\subsection{Main Results}
Following \cite{huang2023opera}, we test our approach using three widely-used metrics, including CHAIR evaluation, GPT-4 assisted evaluation and GPT-4V assisted evaluation.
\begin{table}[t]
    \caption{Evaluation results on CHAIR metric. Smaller values indicate fewer hallucinations.}
    \vspace{0.5em}
    \setlength{\tabcolsep}{1mm}
    \renewcommand{\arraystretch}{0.9}
    \footnotesize
    \centering
    \begin{adjustbox}{width=1.0\columnwidth,center}
    \begin{tabular}{l c c c c c c c c}
        \toprule
        
        & \multicolumn{2}{c}{InstructBLIP} 
        & \multicolumn{2}{c}{MiniGPT-4} 
        & \multicolumn{2}{c}{LLaVA-1.5} 
        & \multicolumn{2}{c}{Shikra} 
        \\
        \cmidrule{2-3}
        \cmidrule{4-5}
        \cmidrule{6-7}
        \cmidrule{8-9}
        Method & $C_S$ & $C_I$ & $C_S$ & $C_I$ & $C_S$ & $C_I$ & $C_S$ & $C_I$
        \\
        \midrule
        Greedy & 30.0 & 14.5 & 24.2 & 8.2 & 20.6 & 6.2 & 22.0 & 7.0
        \\
        Nucleus & 30.4 & 15.7 & 23.6 & 8.3 & 26.2 & 8.5 & 22.6 & 7.6\\
        Beam Search & 21.4 & 7.2 & 23.6 & 7.8 & 18.8 & 5.9 & 20.2 & 6.4
        \\
        \midrule
          ReCaption & 17.2 & 6.6 & 21.8 & 8.1 & 13.8 & 5.0 & 13.9 & 5.5\\
         Woodpecker & 15.5 & 6.5 & 21.5 & 7.8 & 13.4 & 4.8 & 13.6 & \textbf{5.1}\\
         \midrule
        CD & 22.0 & 7.1 & 23.8 & 8.3 & 20.9 & 6.0 & 20.4 & 6.4\\
        DoLa & 22.2 & 7.1 & 24.2 & 8.2 & 20.4 & 6.3 & 20.2 & 6.3
        \\ 
        VCD & 19.7 & 7.0 & 23.7 & 8.0 & 18.9 & 5.7 & 19.5 & 6.2\\
        OPERA & 16.6 & 6.8 & 22.6 & {8.2} & 14.2 & 5.2 & 14.2 & 5.9
        \\
         \midrule
         \rowcolor{mygray} \textbf{IBD} & \textbf{15.0} & \textbf{6.2} & \textbf{21.0} & \textbf{7.4} & \textbf{12.7} & \textbf{4.5} & \textbf{13.2} & 5.2\\
        \bottomrule
    \end{tabular}
    \end{adjustbox}
    \vspace{-0.5em}
    \label{chair_result}
\end{table}

\noindent \textbf{CHAIR Evaluation. }
Caption Hallucination Assessment with Image Relevance (CHAIR) is a commonly-used metric for evaluating the degree of object hallucination in image captioning tasks. It assesses the proportion of objects that are present in the generated captions but absent in the ground truth. CHAIR encompasses two dimensions, denoted as $C_{S}$ and $C_{I}$, which evaluate the degree of object hallucination from sentence-level and image-level perspectives, respectively. $C_{S}$ and $C_{I}$ are computed as:
\begin{equation}
    C_{S} = \frac{|\{\text{\scriptsize captions w/ hallucinated objects}\}|}{|\{\text{\scriptsize all captions}\}|}, C_{I} = \frac{|\{\text{\scriptsize hallucinated objects}\}|}{|\{\text{\scriptsize all mentioned objects}\}|}.
\end{equation}
We follow the same setting as \cite{huang2023opera} by employing the validation set of the COCO 2014 dataset for our evaluation, and using \texttt{Please describe this image in detail.} as LVLM's text input. We present the comparison results in terms of $C_{S}$ and $C_{I}$ in Table.\ref{chair_result}. The methods compared in the table are categorized into three types: (1) the traditional decoding methods, including Greedy Decoding, Nucleus decoding \cite{holtzman2019curious} and Beam Search decoding \cite{boulanger2013audio}; (2) decoding methods specifically designed to address hallucinations, including vanilla CD (CD) \cite{li2022contrastive}, visual CD (VCD) \cite{leng2023mitigating}, DOLA \cite{chuang2023dola} and OPERA \cite{huang2023opera}; and (3) assistance-based methods that employ an additional model to mitigate hallucinations or rewrite descriptions, including ReCaption \cite{wang2023mitigating} and Woodpecker \cite{yin2023woodpecker}. The results show that our IBD can consistently outperform all other types of methods across different baseline models, demonstrating its high effectiveness. Also note that IBD requires only a minimal number (74K) of additional parameters for the prompt $P$ beyond the baseline model. This distinguishes it from other assistance-based methods, which need to use additional LLMs or network architectures with an excessive number of additional parameters. As shown in Table.\ref{comp_para}, compared to these methods, IBD is both more effective and efficient.

\begin{figure}
    \centering
    \includegraphics[width=1\linewidth]{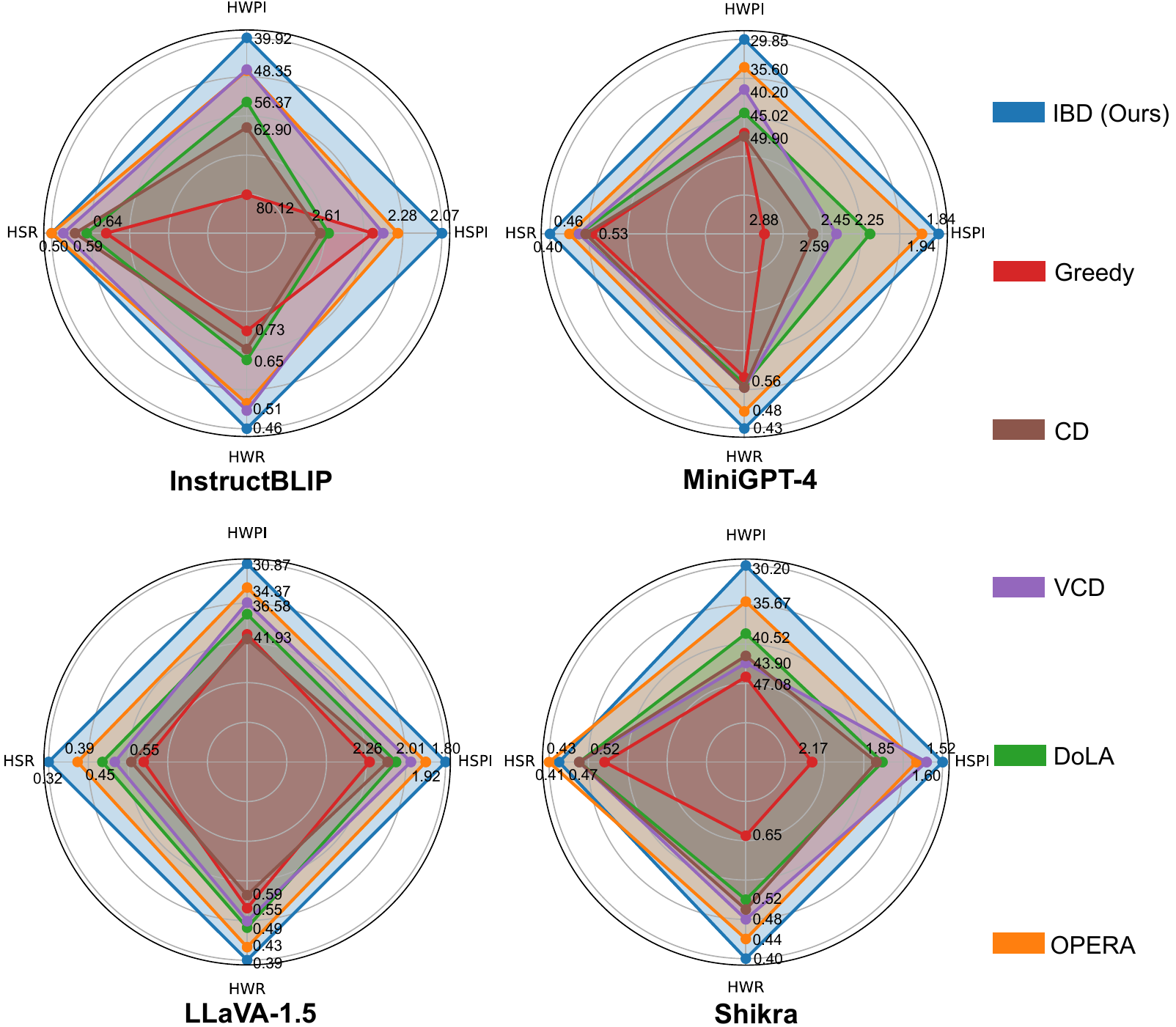}
    \vspace{-0.8\baselineskip}
    \caption{Evaluation results assisted by GPT-4, including 4 metrics: the number of hallucinated sentences per image (HSPI), the number of hallucinated words per image (HWPI), the ratio of hallucinated sentences (HSR), and the ratio of hallucinated words (HWR). Lower values indicate fewer hallucinations.}
    \label{gpt4_metric}
\end{figure}

\begin{table}[t]
     \caption{Comparison with other assistance-based methods in terms of additional parameters/tools and the CHAIR metric. Lower values of $C_{S}$ and $C_{I}$ indicate fewer hallucinations.}
    \footnotesize
    \centering
    \setlength{\tabcolsep}{3pt}
    \renewcommand{\arraystretch}{0.92}
    \begin{adjustbox}{width=1.0\columnwidth,center}
    \begin{tabular}{l c c c c c}
        \toprule
        & & \multicolumn{2}{c}{MiniGPT-4} & \multicolumn{2}{c}{LLaVA-1.5} 
        \\
        \cmidrule{3-4}
        \cmidrule{5-6}
        Method & Additional Param/Tool & $C_{S}$ & $C_{I}$ & $C_{S}$ & $C_{I}$
        \\
        \midrule
        ReCaption & GPT4 & 21.8 & 8.1 & 13.8 & 5.0\\
        WoodPecker & 273M & 21.5 & 7.8 & 13.4 & 4.8\\
        \midrule
        \rowcolor{mygray} \textbf{IBD} & \textbf{68K} & \textbf{21.0} & \textbf{7.4} & \textbf{12.7} & \textbf{4.5}\\
        \bottomrule
    \end{tabular}
    \end{adjustbox}
    \vspace{-1\baselineskip}
    \label{comp_para}
\end{table}

\noindent \textbf{GPT-4 Assisted Evaluation. }A limitation of CHAIR is that it can only assess object-existence-level hallucination. To further measure hallucinations at other levels, such as attributes and relations, we follow the method in \cite{huang2023opera} by employing GPT-4 to evaluate the discrepancies between the descriptions generated by the LVLM and the ground truth descriptions on the VG dataset \cite{krishna2017visual}, thus assessing the degree of hallucinations in a more comprehensive manner. For a fair comparison, we prompt GPT-4 using the same method as \cite{huang2023opera}, asking GPT4 to produce the assessment for each description with six metrics. Readers can refer to Sec.4.4 of \cite{huang2023opera} for detailed explanations of each metric. We conduct comparisons on four metrics that are highly correlated to the hallucination: the number of hallucinated sentences per image (HSPI), the number of hallucinated words per image (HWPI), the ratio of hallucinated sentences (HSR), and the ratio of hallucinated words (HWR). The results of these comparisons are presented in Fig.\ref{gpt4_metric}. Across these metrics, our method demonstrates a significant advantage, which indicates the high effectiveness of our method in mitigating LVLM's hallucinations.

\begin{table}[t]
    \caption{Evaluation results assisted by GPT-4V, including two metrics: correctness ($C$) and detailedness ($D$). Higher values indicate fewer hallucinations.}
    \vspace{0.5em}
    \footnotesize
    \centering
    \setlength{\tabcolsep}{1.5mm}
    \renewcommand{\arraystretch}{1}
    \begin{adjustbox}{width=1.0\columnwidth,center}
    \begin{tabular}{l c c c c c c c c}
        \toprule
        & \multicolumn{2}{c}{InstructBLIP} 
        & \multicolumn{2}{c}{MiniGPT-4} 
        & \multicolumn{2}{c}{LLaVA-1.5} 
        & \multicolumn{2}{c}{Shikra} 
        \\
        \cmidrule{2-3}
        \cmidrule{4-5}
        \cmidrule{6-7}
        \cmidrule{8-9}
        Method & $C$ & $D$ & $C$ & $D$ & $C$ & $D$ & $C$ & $D$
        \\
        \midrule
        Beam Search & 5.4 & 5.2 & 4.5 & 4.9 & 5.9 & 5.2 & 5.2 & 5.0 \\
        CD & 5.4 & 5.4 & 4.8 & 5.2 & 6.2 & 5.2 & 5.4 & 5.2 \\
        OPERA & 6.0 & 5.6 & 5.2 & 5.0 & 6.3 & 4.9 & 5.9 & 5.0 \\
        VCD & 6.0 & 5.8 &  5.5 & 5.2 & 6.4 & 5.0 &  6.1 & 5.1 \\
        DoLa & 5.8 & 5.9 & 5.0 & 5.0 & 6.3 & 5.3 & 5.8 & 5.2\\
        \midrule
        \rowcolor{mygray} \textbf{IBD} & \textbf{6.3} & \textbf{6.1} & \textbf{5.8} & \textbf{5.4} & \textbf{6.7} & \textbf{5.5} & \textbf{6.6} & \textbf{5.8}\\
        \bottomrule
    \end{tabular}
    \end{adjustbox}
    \label{gpt4v_metric}
\vspace{-1\baselineskip}
\end{table}

\noindent \textbf{GPT-4V Assisted Evaluation. }We further evaluate LVLM's description quality on the MSCOCO dataset by inputting both the image and the generated description into GPT-4V. Concretely, utilizing the method proposed by \cite{huang2023opera}, we instruct GPT-4V to score these descriptions on two metrics, Accuracy and Detailedness, on a scale from 0 to 10. The results are presented in Table.\ref{gpt4v_metric}. In comparison to other methods, our method achieves a significant enhancement. Particularly on the Shikra baseline model, our method surpasses the runner-up by 11.5\% in the Detailedness metric. Such gains can be attributed to our method's enhanced exploitation of image information, which enables the generation of descriptions that are more comprehensive with fewer information omissions.

\subsection{Ablation Study}\label{ablation}
\noindent \textbf{Evaluation of Different Components. }We conduct an ablation study based on the CHAIR metric to examine the effectiveness of each design component. Specifically, as shown in Table.\ref{ablation_components}, the evaluated components include: (1) dynamic adjustment (DA) for decoding (Sec.\ref{dynamic_adjustment}), (2) prompt $P$ for finetuning (Sec.\ref{dynamic_adjustment}); and (3) the adaptive plausibility constraint (APC) as in Eq.\ref{full} and Eq.\ref{condition}. Additionally, we evaluate two indicators $I_{sim}$ and $I_{con}$ used within the dynamic adjustment method (Eq.\ref{dynamic}). Removing any of these components would result in a notable decrease in performance. In Eq.\ref{i_fun}, we compute $I_{con}$ by using the distances between predictions from different layers, rather than directly employing a POS tagging tool, which may suffer from semantic ambiguities (details in sec.\ref{dynamic_adjustment}). Our experiments confirm this notion, with the use of a POS-based $I_{con}$ tagged by spacy \cite{spacy2} significantly degrading performance. These results demonstrate the rationality and effectiveness of different design components in our method. 

\begin{table}[t]
    \caption{Evaluation of IBD's different components on CHAIR metric. 
    DA and APC refer to dynamic adjustment (Sec.\ref{dynamic_adjustment}) and adaptive plausibility constraint (Sec.\ref{full_method}) respectively. }
    \vspace{0.5em}
    \setlength{\tabcolsep}{1mm}
    \footnotesize
    \centering
    \begin{adjustbox}{width=1.0\columnwidth,center}
    \begin{tabular}{l c c c c c c c c}
        \toprule
        \vspace{0.5em}
        & \multicolumn{2}{c}{InstructBLIP} 
        & \multicolumn{2}{c}{MiniGPT-4} 
        & \multicolumn{2}{c}{LLaVA-1.5} 
        & \multicolumn{2}{c}{Shikra} 
        \\
        \cmidrule{2-3}
        \cmidrule{4-5}
        \cmidrule{6-7}
        \cmidrule{8-9}
        Method & $C_S$ & $C_I$ & $C_S$ & $C_I$ & $C_S$ & $C_I$ & $C_S$ & $C_I$
        \\
        \midrule
        \rowcolor{mygray} \textbf{IBD} & \textbf{15.0} & \textbf{6.2} & \textbf{21.0} & \textbf{7.4} & \textbf{12.7} & \textbf{4.5} & \textbf{13.2} & \textbf{5.2}\\
        \midrule
        w/o DA & 19.5 & 7.0 & 23.3 & 8.0 & 18.0 & 5.5 & 18.9 & 5.9\\
        w/o Prompt $P$ & 17.2 & 6.5 & 22.0 & 7.8 & 15.8 & 5.0 & 16.4 & 5.6\\
        w/o APC & 15.8 & 6.4 & 22.0 & 7.7 & 13.5 & 4.6 & 14.1 & 5.5\\
        \midrule
        w/o $I_{sim}$ in Eq.\ref{dynamic} & 15.7 & 6.4 & 21.8 & 7.6 & 13.6 & 4.8 & 13.9 & 5.5\\
        w/o $I_{con}$ in Eq.\ref{dynamic} & 18.3 & 6.8 & 22.5 & 7.9 & 16.5 & 5.1 & 18.0 & 5.8\\
        \midrule
        w/ POS-based $I_{con}$ & 16.7 & 6.5 & 21.8 & 7.7 & 14.5 & 4.8 & 15.7 & 5.6\\
        \bottomrule
    \end{tabular}
    \end{adjustbox}
    \label{ablation_components}
\end{table}

\begin{figure}
    \centering
    \includegraphics[width=1\linewidth]{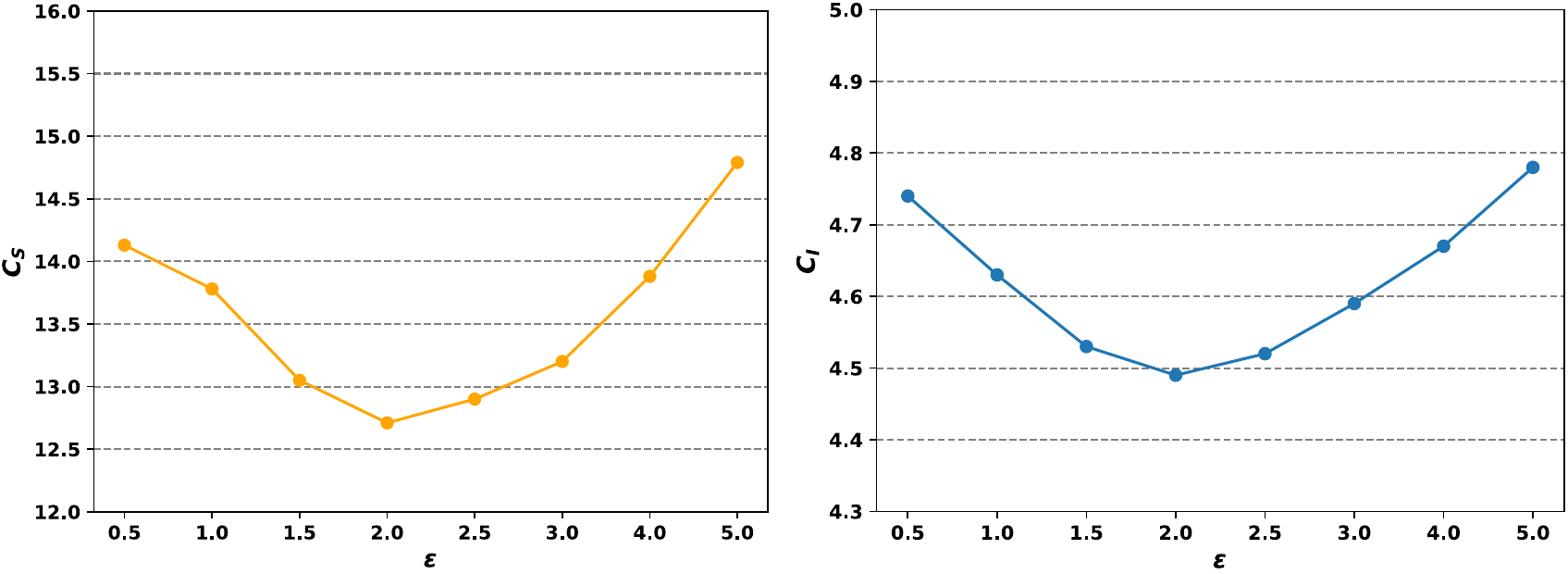}
    \vspace{-1.5\baselineskip}
    \caption{Evaluation results when using different $\epsilon$. The left figure shows the results of the $C_{S}$ metric in CHAIR, and right figure shows the results of the $C_{I}$ metric in CHAIR. Smaller values indicate fewer hallucinations.}
    \label{epsilon_ablation}
    \vspace{-1\baselineskip}
\end{figure}

\noindent \textbf{Settings for Hyper-parameter $\epsilon$. }
As shown in Eq.\ref{weight}, we introduce a factor added to the QK-multiplication result to bias model $\hat{\theta}$ towards image information. This factor is equal to a hyper-parameter $\epsilon$ when the key corresponds to an image token. In Fig.\ref{epsilon_ablation}, using the CHAIR metric, we present the performance when using different values of $\epsilon$. The results indicate suboptimal performance when $\epsilon$ is either too small or too large. This is attributed to the fact that an excessively small $\epsilon$ results in an insufficient bias towards the image, while an overly large $\epsilon$ introduces too much noise, thus impairing the model's performance. When $1.5<\epsilon<3$, the performance can keep relatively stable.

\vspace{-0.25\baselineskip}
\section{Conclusion}
\vspace{-0.25\baselineskip}
This paper proposes image-biased decoding (IBD), a novel decoding method that aims at alleviating the issue of hallucinations in LVLMs. Our approach involves conducting a prediction contrast between the original model and an image-biased model to amplify the accurate information associated with image content, thereby improving the factuality of the generated text. In addition, we design a dynamic adjustment strategy that flexibly handles different types of vocabulary. Experimental results demonstrate the high effectiveness of IBD in addressing LVLM's hallucination challenges. We hope our innovative method and detailed statistical analysis can provide valuable insights for future research in this field.

\subsection*{Impact Statement}
This paper presents work whose goal is to advance the field of Machine Learning. There are many potential societal consequences of our work, none which we feel must be specifically highlighted here.
\bibliography{arxiv}
\bibliographystyle{icml2024}

\newpage
\appendix
\onecolumn
\section{More Analyses}
\begin{figure}[t]
    \centering
    \subfigure[Content Words]{
        \label{content_dola}
        \centering
        \includegraphics[width=0.3\linewidth]{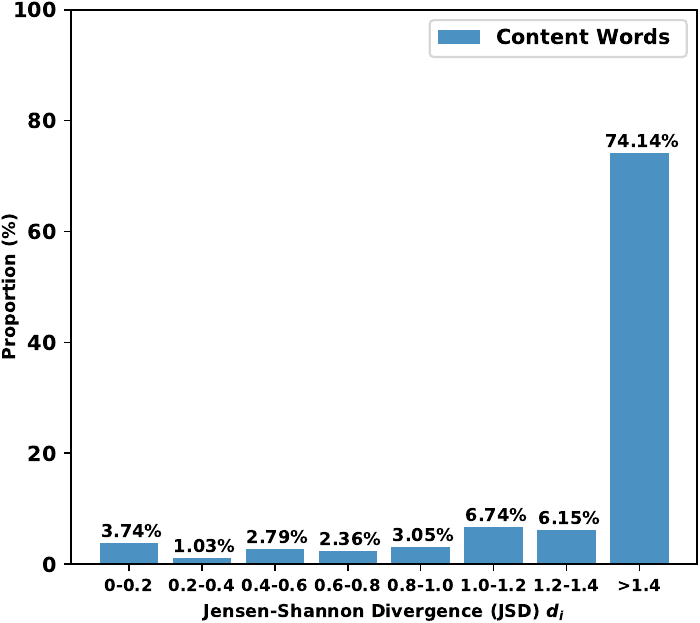}
        }
    \centering
    \subfigure[Function Words]{
        \label{function_dola}
        \centering
        \includegraphics[width=0.3\linewidth]{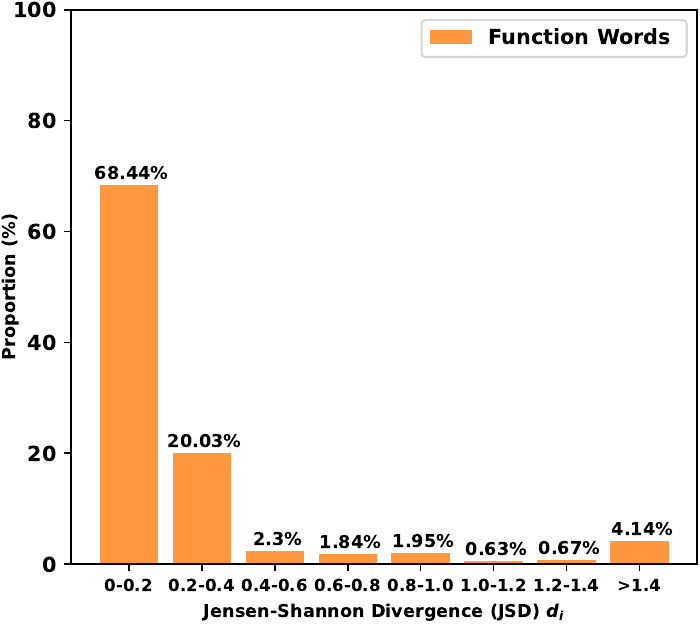}
        }
    \centering
     \subfigure[Content Subwords]{
        \label{subword_dola}
        \centering
        \includegraphics[width=0.3\linewidth]{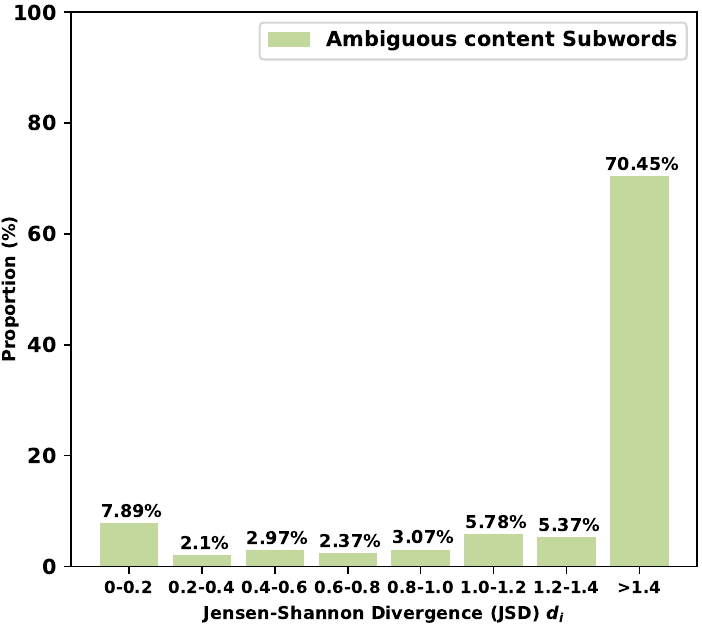}
        }
    \caption{(a), (b) and (c) respectively present the statistical results for content words, function words and content subwords in the COCO Caption dataset. X-axis represents the range of JSD (Jensen-Shannon divergence), Y-axis represents the proportion of conditions where the computed JSD falls into each range.}
    \label{fig_validate_dola}
\end{figure}
\subsection{Early Exit's Patterns for Different Words}
Prior research \cite{chuang2023dola} has shown that when employing an early exit strategy in language models to predict function words, the LLM tends to determine the token to generate within the middle layers, and keeping the predictions almost unchanged in the subsequent higher layers. In contrast, when predicting content words that are more complex, the LLM continues to change its predictions in the last few layers. We leverage these observed patterns to compute the indicator $I_{con}$ as in Eq.\ref{i_fun}. To ensure the reliability of our method, we have conducted the following two validations:

(1) We first examine whether the patterns observed in LLMs are also applicable to LVLMs. Using the COCO caption dataset, we apply Eq.\ref{i_fun} to calculate the Jensen-Shannon Distance (JSD) between the predictions of the 24th layer and the final layer of LLaVA-1.5 at each time step. We divide the results into two parts based on whether the predicted token is a content word token or a function word token, and then separately analyze the distributions of JSD as shown in Fig.\ref{content_dola} and Fig.\ref{function_dola}. We observe that 74.14\% of the time, the JSD is greater than 1.4 when predicting content words, indicating that the LVLM is still significantly changing its predictions in the deeper layers. In contrast, 88.47\% of the time the JSD is less than 0.4 when predicting function words, which suggests that the network has already decided on the predictions for most of the function words in the shallower layers and maintained them unchanged in subsequent layers. These results are consistent with the patterns observed for LLMs as reported in \cite{chuang2023dola}, validating that these patterns regarding the difference between function words and content words remain applicable in the context of LVLMs.

(2) As discussed in Sec.\ref{dynamic_adjustment}, some subwords of content words may possess semantic ambiguity. For instance, without seeing the complete content word `onion', its subword `on' could be mistakenly identified as a function word. To ensure the robustness of our IBD technique when processing such words, we employ a method similar to (1), using the COCO caption dataset and Jensen-Shannon Divergence (JSD) to verify whether the aforementioned patterns are also applicable to content subwords with ambiguity. As shown in Fig.\ref{subword_dola}, we observe that 70.45\% of the time, the JSD is greater than 1.4 when predicting ambiguous content subwords, indicating that the predictions still change in LVLM's deeper layers. These findings suggest that the ambiguous content subwords also follow the patterns of content words discovered by \cite{chuang2023dola}.

\subsection{Image-biased Hallucinations in Mainstream Evaluation Frameworks}
In this section, we employ GPT-4V to investigate a phenomenon we previously noted in Sec.\ref{discuss_image_biased}: image-biased hallucinations rarely occur within datasets of mainstream hallucination evaluation tasks. More specifically, we utilize a comprehensive evaluation dataset, MME \cite{fu2023mme}, which encompasses 14 distinct subtasks, to conduct tests with LLaVA-1.5. We randomly select 500 instances from the set of questions that LLaVA-1.5 incorrectly answers, and ask GPT-4V to assist in determining whether the error could potentially be attributed to image-biased hallucinations using the following prompts:

\texttt{I will give you an image, a question related to that image, a correct response, and an erroneous response generated by a model. You are required to determine whether the model's erroneous response is due to image-biased hallucinations. Here, hallucinations refer to the model generating responses inconsistent or incorrect with the input image and question. Image-biased hallucinations refer to the phenomenon that the model generates incorrect responses due to being influenced by certain regions or information in the image. For example, if an image shows a soccer star with a table tennis racket nearby, and the question is whether the person in the image is a table tennis player, the model might incorrectly respond yes due to the presence of the table tennis racket in the image, which is an example of image-biased hallucinations. Your output should be: Judgment (erroneous response generated by the model is or is not due to image-biased hallucinations ); <reason>.}

We direct GPT-4v to evaluate all 500 instances using the above prompts. The results reveal that only in 4\% of instances did GPT-4v attribute the erroneous responses generated by the model to image-biased hallucinations. This finding demonstrates the viewpoint we presented in Sec.\ref{discuss_image_biased}, indicating that image-biased hallucinations rarely occur in the mainstream frameworks used for evaluating LVLMs.

\vspace{-0.5\baselineskip}
\section{More Discussions: Why not directly train an image-biased LVLM?}
\begin{wraptable}{r}{.5\textwidth}{
    \renewcommand{\arraystretch}{0.85}
    \setlength\tabcolsep{10pt}
    \begin{tabular}{l | c c}
    \toprule
    Method & $C_{S}$ & $C_{I}$ \\
    \midrule
    Original LVLM $\theta$ & 20.61 & 6.18\\
    \midrule
    Image-biased LVLM $\hat{\theta}$ ($\epsilon=0.5$) & 20.45 & 6.13\\
    Image-biased LVLM $\hat{\theta}$ ($\epsilon=1.0$)  & 20.28 & 6.03\\
      Image-biased LVLM $\hat{\theta}$ ($\epsilon=2.0$)  & 20.25 & 6.06\\
        Image-biased LVLM $\hat{\theta}$ ($\epsilon=3.0$)  & 20.74 & 6.24\\
         Image-biased LVLM $\hat{\theta}$ ($\epsilon=4.0$)  & 21.90 & 6.58\\
          Image-biased LVLM $\hat{\theta}$ ($\epsilon=5.0$)  & 23.45 & 7.19\\
     \bottomrule
    \end{tabular}
    \vspace{-0.5\baselineskip}
    \caption{Evaluation results based on LLaVA-1.5. Smaller values indicate fewer hallucinations.}
    \label{direct_train}
}
\end{wraptable}
IBD derives the next-token probability distribution by contrasting predictions from a conventional LVLM with those of an image-biased LVLM, thereby amplifying the correct information highly correlated with image content while mitigating the hallucinatory errors caused by excessive dependence on text. An intuitive question that naturally arises is whether we can alleviate the issue of text-biased hallucinations by directly training an image-biased model without the need for using contrastive decoding techniques. However, when we explored this approach, the results did not meet our expectations. Specifically, we follow the method illustrated in Sec.\ref{sec:image_biased} to construct an image-biased model $\hat{\theta}$ and finetune it using the LLaVA-Instruct-150K dataset. The performance of employing different $\epsilon$ in Eq.\ref{weight} to construct $\hat{\theta}$, which determines the extent of image bias, are presented in Table.\ref{direct_train}. It is observed that $\hat{\theta}$ can exhibit marginal improvements when $0.5<\epsilon<2$ but these enhancements are not significant. While as $\epsilon$ is set higher than 3, $\hat{\theta}$'s performance even becomes worse than $\theta$.

We interpret the phenomena observed as follows: when $\epsilon$ is small, changes in $\hat{\theta}$'s prediction compared to the original model $\theta$ is minimal. As a consequence, while the predictive probability of the ground truth token may increase and that of the hallucinatory token may decrease, the minor degree of change may still be insufficient for the ground truth token's probability to surpass that of the hallucinated token. As a result, the hallucinatory token is still likely to be selected as the predicted outcome, resulting in no significant improvement in performance. Conversely, when $\epsilon$ is excessively large, the substantial alterations in the model structure may inadvertently lead to a degradation in performance. Consequently, both excessively small and large values of $\epsilon$ prove inappropriate for the model, and thus the performance of $\hat{\theta}$ consistently fails to reach satisfactory levels under various settings of $\epsilon$. In contrast, our image-biased decoding method based on contrastive techniques focuses on the variation in predictive probabilities from $\theta$ to $\hat{\theta}$ rather than the absolute predictions of $\hat{\theta}$ alone, thus it can more effectively identify the correct token without text-biased hallucinations using a small value of $\epsilon$.

\begin{figure}[t]
    \centering
    \vspace{-0.5\baselineskip}

    \subfigure[$\alpha$]{
        \label{alpha_val}
        \centering
        \includegraphics[width=0.3\linewidth]{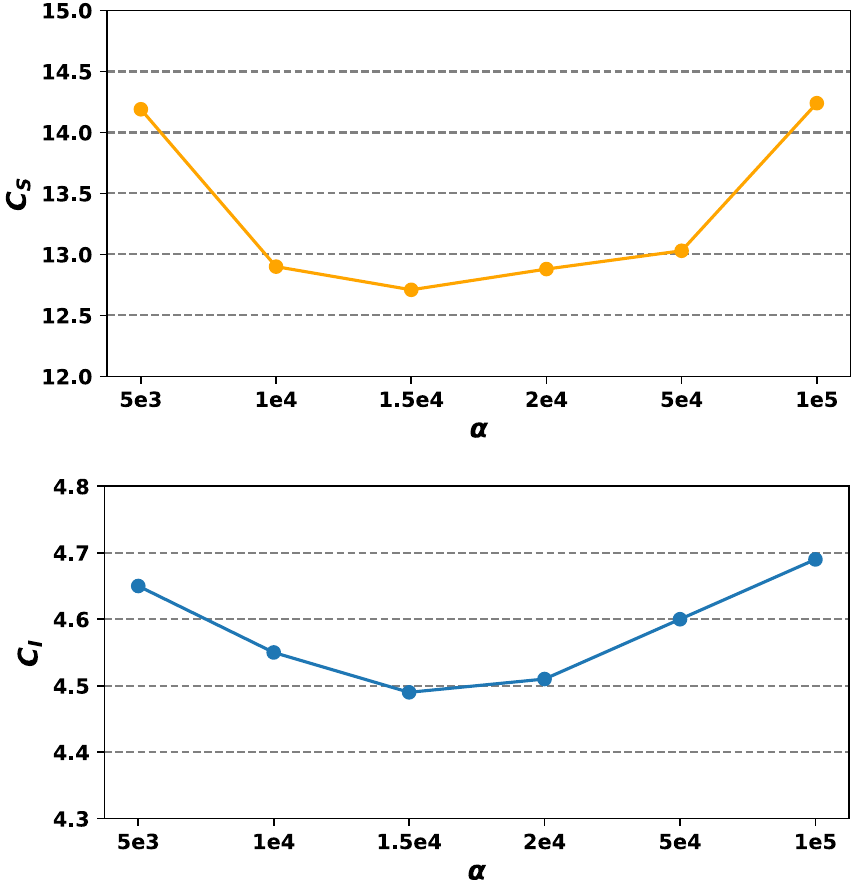}
        }
    \subfigure[$\beta$]{
        \label{beta_val}
        \centering
        \includegraphics[width=0.3\linewidth]{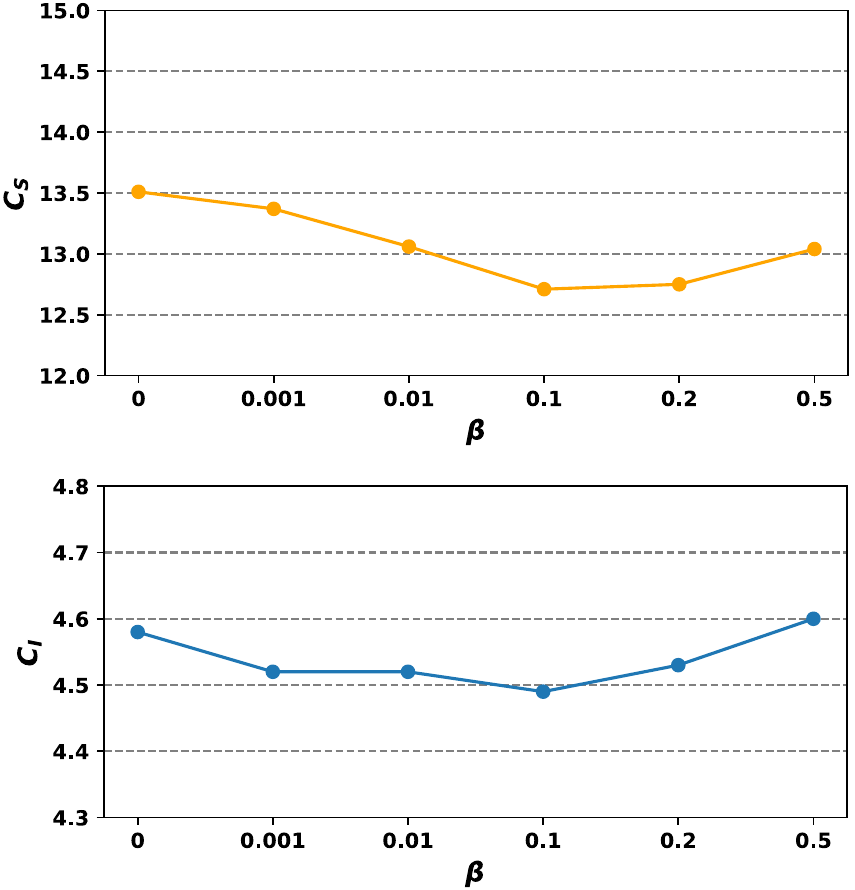}
        }
    \subfigure[Layer]{
        \label{layer_val}
        \centering
        \includegraphics[width=0.3\linewidth]{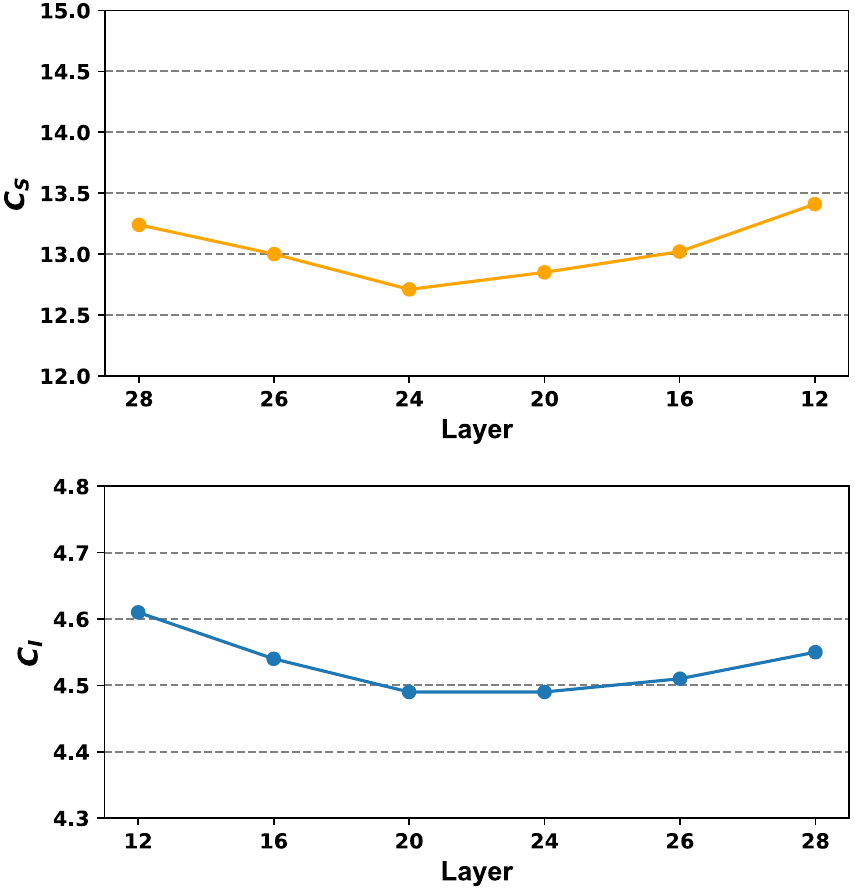}
        }
    \vspace{-0.5\baselineskip}
    \caption{Evaluation results when using different $\alpha$, $\beta$ and different layers where we get the early-exiting prediction $\widetilde{p}_{\theta}\left(y_{i}|v, t, y_{<i}\right)$ for computing $I_{con}$ in Eq.\ref{i_fun}. Figures in the first row show the results of the $C_{S}$ metric in CHAIR, and figures in the second row show the results of the $C_{I}$ metric in CHAIR. Smaller values indicate fewer hallucinations.}
    \label{aaa}
\end{figure}

\vspace{-0.5\baselineskip}
\section{More Ablation Study for Hyper-parameters}
In addition to $\epsilon$ evaluated in Sec.\ref{ablation} of the main paper, based on LLaVA-1.5 baseline and CHAIR metric, we further validate three other hyper-parameters used in IBD -- $\alpha$ in Eq.\ref{dynamic}, $\beta$ in Eq.\ref{condition} and the layer where we get the early-exiting prediction $\widetilde{p}_{\theta}\left(y_{i}|v, t, y_{<i}\right)$ for computing $I_{con}$ in Eq.\ref{i_fun}. The results of the assessment regarding these hyper-parameters are presented in Fig.\ref{alpha_val}, Fig.\ref{beta_val} and Fig.\ref{layer_val}, respectively. The consistent performance obtained with varying hyper-parameter values demonstrates the high robustness of our proposed method.

\begin{figure}[t]
    \centering
    \includegraphics[width=1\linewidth]{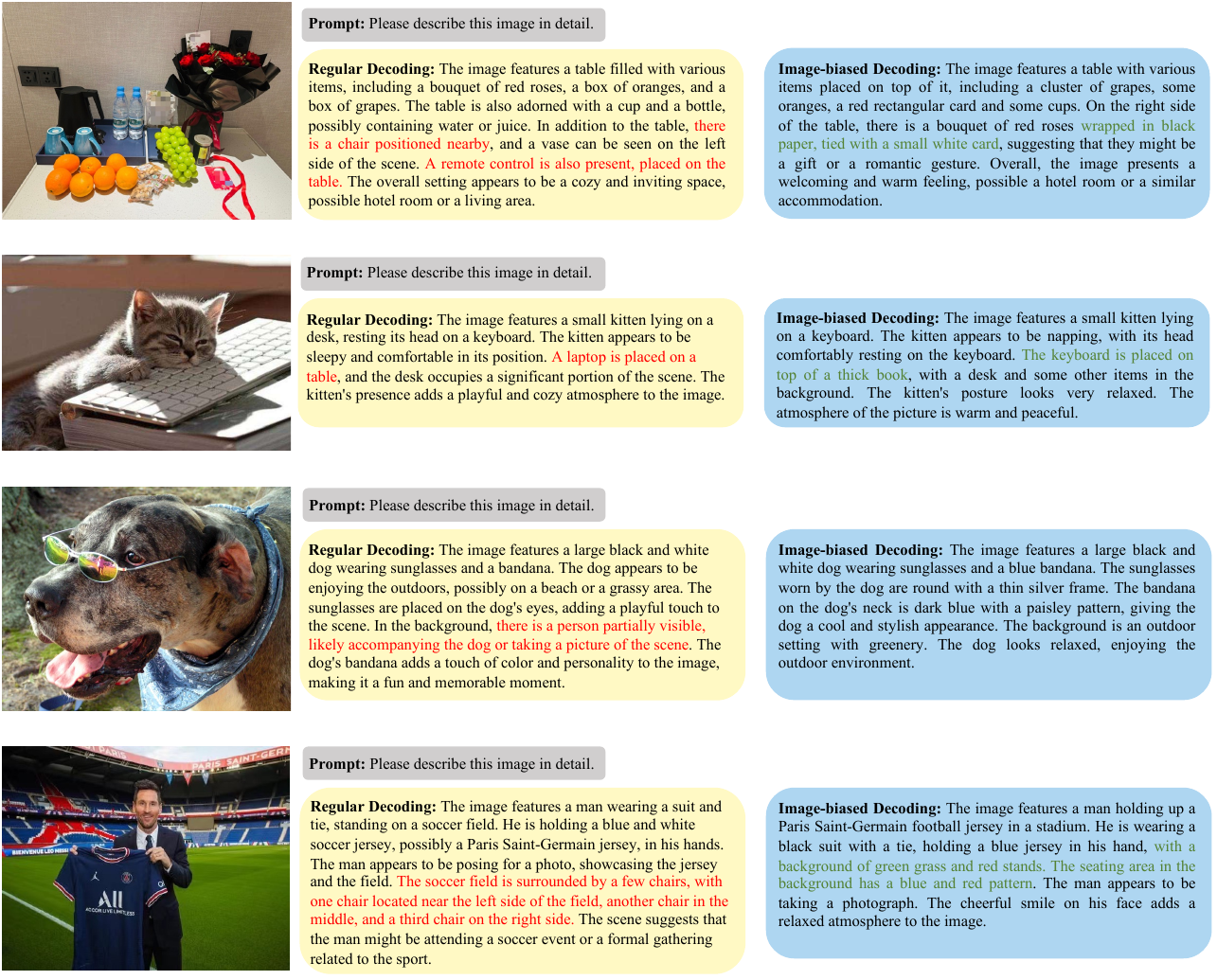}
    \vspace{-1\baselineskip}
    \caption{Examples to show the comparison between the regular maximum likelihood decoding and our image-biased decoding (IBD) based on LLaVA-1.5. Hallucinations generated from LVLM's regular decoding are highlighted in {\color{red} red}. Information omitted during regular decoding but generated by IBD is highlighted in {\color{green} green}.}
    \label{example}
\vspace{-0.5\baselineskip}
\end{figure}

\end{document}